\documentclass[journal,comsoc]{IEEEtran}
\usepackage[T1]{fontenc}% optional T1 font encoding

\ifCLASSINFOpdf
\else
\fi
\usepackage{amsmath}
\usepackage[cmintegrals]{newtxmath}
\usepackage{graphicx}
\usepackage{multirow}
\usepackage{arydshln}
\usepackage{gensymb}
\usepackage{amsmath}
\usepackage{subcaption}
\usepackage{pgfplots}
\usepackage[linesnumbered,ruled,vlined]{algorithm2e}
\newcommand\norm[1]{\left\lVert#1\right\rVert}

\DeclareMathOperator*{\argmin}{arg\,min}

\begin{document}
\title{A Unified Framework for Speech Separation}

\author{Fahimeh~Bahmaninezhad,~\IEEEmembership{}
        Shi-Xiong~Zhang,~\IEEEmembership{}
        Yong~Xu,~\IEEEmembership{}
        Meng~Yu,~\IEEEmembership{}
        John H.L.~Hansen,~\IEEEmembership{}
        and~Dong~Yu~\IEEEmembership{}% <-this % stops a space
\thanks{F. Bahmaninezhad and J. H.L. Hansen are with the Department of Electrical and Computer Engineering, University of Texas at Dallas, Richardson, TX, 75080 USA e-mail: \{fahimeh.bahmaninezhad, john.hansen\}@utdallas.edu.}% <-this % stops a space

\thanks{SX. Zhang, Y. Xu, M. Yu, and D. Yu are with Tencent AI Lab, Bellevue WA, 98004 USA email: \{auszhang,lucayongxu,raymondmyu,dyu\}@tencent.com .}% <-this % stops a space
\thanks{Manuscript received April 19, 2005; revised August 26, 2015.}}

\maketitle

\begin{abstract}

Speech separation refers to extracting each individual speech source in a given mixed signal. Recent advancements in speech separation and ongoing research in this area, have made these approaches as promising techniques for pre-processing naturalistic audio streams.
After incorporating deep learning techniques into speech separation, performance on these systems is improving faster. 
The initial solutions introduced for deep-learning based speech separation analyzed the speech signals into time-frequency domain with STFT (short-time Fourier transform), specifically; and then encoded mixed signals were fed into a deep neural network based separator. Most recently, new methods are introduced to separate waveform of the mixed signal directly without analyzing them using STFT.
Here, we introduce a unified framework to include both spectrogram and waveform separations into a single structure, while being only different in the kernel function used to encode and decode the data; where, both can achieve competitive performance. 
This new framework provides flexibility; in addition, depending on the characteristics of the data, or limitations of the memory and latency can set the hyper-parameters to flow in a pipeline of the framework which fits the task properly. We extend single-channel speech separation into multi-channel framework with end-to-end training of the network while optimizing the speech separation criterion (i.e., Si-SNR) directly.
We emphasize on how tied kernel functions for calculating spatial features, encoder, and decoder in multi-channel framework can be effective.
We simulate spatialized reverberate data for both WSJ0 and LibriSpeech corpora here, and while these two sets of data are different in the matter of size and duration, the effect of capturing shorter and longer dependencies of previous/+future samples are studied in detail. We report SDR (source-to-distortion ratio), Si-SNR (scale-invariant source-to-noise ratio) and PESQ (perceptual evaluation of speech quality) criteria to evaluate the performance of developed solutions objectively and subjectively. 
\end{abstract}


% \fi
\IEEEpeerreviewmaketitle

\section{Introduction}
\label{sec:intro}

\IEEEPARstart{T}{he} cocktail party problem \cite{mcdermott2009cocktail} or dinner party problem \cite{barker2018fifth} is about complex real-world environments with conversational speech signals overlapped with other interfering speech, noise or music sources and how to focus the auditory attentions towards only one of these sources \cite{barker2018fifth}. 
Robust speech processing (e.g., speaker or speech recognition) of the data collected from this environment is becoming more of an interest over the past decade with growing voice assistant devices such as Amazon Alexa or Google Home \cite{kim2017generation, li2017acoustic, kim2018sound}.
In addition, speech processing of data collected in controlled laboratory conditions has improved enough to not be a huge concern anymore \cite{settle2018end, fu2018end}.
Multiple challenges have been organized before as well \cite{barker2018fifth, cooke2010monaural, barker2015third} with focusing on the dinner party problem, which all emphasize on the importance of improving speech processing techniques for the naturalistic audio streams.
One solution to process this data is to first separate all overlapping sources and then process each of them separately \cite{settle2018end}. The task of extracting all overlapping speech sources in a given mixed speech signal refers to the \textit{speech separation} \cite{kolbaek2017multitalker, huang2014deep, wang2018supervised}.
Speech separation is a special scenario of source separation problem \cite{vincent2018audio, araki2007underdetermined, lee2005blind, venkataramani2018end}; where the focus is only on the overlapping speech signal sources and other interferences such as music or noise signals are not the main concern of the study.

Monaural speech separation \cite{kolbaek2017multitalker, yu2017permutation} is based on masking the time-frequency (TF) representation of the signal. The initial solutions introduced for TF-masking of the given mixed signal were using heuristic, knowledge-based information or matrix factorization techniques to cluster the TF bins of each speaker \cite{cooke2010monaural, schmidt2006single}, and later improved with incorporating statistical models \cite{virtanen2006speech}.
However, these solutions were not effective enough for single-channel speech separation. The drawbacks of the above-mentioned techniques are related to each of the following categories: (i) they are either speaker-dependent \cite{virtanen2006speech}, (ii) or it is difficult to expand the task to any number of overlapping speakers \cite{weng2015deep}. 

The advancements of deep learning techniques had also affected speech separation task to achieve a promising performance after all.
Deep clustering (DC) \cite{hershey2016deep, isik2016single, wang2018alternative} as the first successful solution based on deep learning for TF-masking (the spectrogram of the mixed signal) achieved a great performance in separating overlapped speakers.
DC models embeddings for each TF-bin with a recurrent neural network (RNN) \cite{mikolov2010recurrent}, then the embeddings were clustered into $S$ speaker clusters with a simple clustering technique such as k-means \cite{hershey2016deep}. Performance gained with DC improved further with tuning the hyper-parameters of the network \cite{isik2016single}. Afterwards, new techniques and methods were introduced \cite{kolbaek2017multitalker, yu2017permutation,  wang2018alternative, chen2017deep, williamson2016complex, erdogan2015phase, wang2018end, xu2018single} and all helped to gain more improvement, address the disadvantages of previous methods and initiated new challenges into the speech separation area step-by-step.
For example, the initial solutions proposed for speech separation were designed to work on frame-level or meta-frame level inputs \cite{yu2017permutation}; it arises the problem for speaker tracking when concatenating the consecutive (meta-) frames together \cite{kolbaek2017multitalker, hershey2016deep}. 
Utterance level permutation invariant training (uPIT) \cite{kolbaek2017multitalker} proposed a solution for that. In another topic, reconstructing waveform with phase of the input mixed signal as well can affect the separation performance, which was partially addressed in \cite{wang2018end}.
These research studies, all helped to achieve a single-step, end-to-end training for frequency-domain single-channel speech separation, which is more simple and performs better than the initial proposed solution in \cite{hershey2016deep}.

All methods mentioned above are applied in time-frequency domain; i.e., the input mixed signal was first analyzed into magnitude and phase components with STFT (short-time Fourier transform), and only the magnitude component was used to train the separator model. Finally, the phase of the mixed signal (or an improved phase component) was used in addition to the separated magnitude component to reconstruct the output time-domain waveform using ISTFT (inverse short-time Fourier transform) for each speaker.
Although, the reported results on the frequency domain models are promising, inherently these models have disadvantages which limit their capabilities. The constraints of the the frequency-domain models include: (i) because the phase usually is directly copied from the mixed signal, at reconstruction the performance degrades mainly because of the phase. However, there are studies to improve the phase component \cite{wang2018end} or using complex-masking \cite{williamson2016complex}; still the proposed solutions are not sufficient enough. (ii) STFT requires almost long temporal window for analysis which can bring limitations into some applications  \cite{luo2018tasnet, luo2019conv}.

More recently, \cite{luo2018tasnet, luo2019conv} introduced solutions for speech separation in time-domain; i.e., the time-domain waveform representation of the mixed signal was down-sampled using 1-D convolution (Conv-1D) layer with a small stride and then was fed into the separation network. At the reconstruction layer as well, the separated outputs were up-sampled with 1-D convolution-transpose (ConvTranspose-1D) layer. The reported results outperformed all previous time-frequency models. However, it is worth mentioning that besides the encoding and decoding strategy, new network structure and loss function were also adopted to gain that improvement.
Here, we propose a unified framework for time- and frequency-domain speech separation with end-to-end training; emphasizing and studying the dis/advantages brought by each of them. We show that frequency-domain separation (even using the phase of the mixed signal) can achieve competitive and in some scenarios superior performance with comprehensive experiments.
Throughout this paper, we refer to time-frequency (or just frequency-domain) speech separation methods as spectrogram separation and time-domain separation as waveform separation.

Speech separation can be an effective module for robust speech processing of real-world data; i.e., data recorded with microphone arrays in reverberate environment with interferences of other speakers, noise and music sources. Most of previous speech separation studies focused on single-channel data. To the best of our knowledge, there is very limited previous studies on speech separation specifically for multi-channel recordings, including \cite{lorrymch, lorrytgt, fbinter}.
(However, there are studies on multi-modal speech separation including image \cite{ephrat2018looking} or multi-channel speech separation for speech recognition \cite{weng2015deep, yoshioka2018multi, chen2018multi}; here, we focus on audio only multi-channel speech separation).
Therefore, in this paper we expand our studies on multi-channel recordings with simulating data not only based on WSJ0 but LibriSpeech as well. In LibriSpeech there are longer duration utterances; more training data is included also, which we believe it can help to make more general and concrete conclusions for the effectiveness of the training loss, encoder/decoder and network structure, etc on speech separation performance.

In this paper, we study multi-channel speech separation for two simulated spatialized reverberate corpora, WSJ0 and LibriSpeech. The novelties offered in this paper can be summarized as:
\begin{itemize}
    \item unified framework for spectrogram and waveform speech separation with single/multi-channel structures.
    \item end-to-end training of multi-channel framework.
    \item controlling latency with very small degradation in the performance.
    \item comprehensive comparison of spectrogram and waveform separation on different data and network configurations.
\end{itemize}

The rest of this paper is organized as follows. In section~\ref{sec:problem} we formulate the monaural single-channel speech separation problem. Next, in section~\ref{sec:unified} we introduce our proposed unified framework, including the multi-channel structure and end-to-end training of our proposed structure. In section~\ref{sec:latency} we study the latency of our solutions and the method proposed to control the latency further.
Section~\ref{sec:exp} introduces our data, the metrics used to evaluate the performance of our developed solutions, experimental setup, results and discussion on the proposed solutions. Finally, conclusion and future work are covered in section~\ref{sec:con}.

\section{Single-Channel Speech Separation}
\label{sec:problem}

Speech separation refers to the task of extracting all overlapping speech sources in a given mixed speech signal.
The monaural speech separation is traditionally defined for the spectrogram separation based solutions \cite{kolbaek2017multitalker}. Here, we present the equations for frequency-domain and in the next section, introduce how that can be extended in time-domain.

For a given linearly mixed single-channel signal $y[n]$, monaural speech separation extracts each individual source signal $x_s[n], s=1...S$ as: 
\begin{equation}
y[n] = \sum_{s=1}^S{x_s[n]}.
\end{equation}

In the time-frequency masking (TF-masking) based solutions \cite{kolbaek2017multitalker}, the estimated $\hat{x}_s$ for the ground-truth $x_s$ is derived based on the following equation,
\begin{equation}
\label{eq2}
|\hat{X}_s(t,f)| = \hat{M}_s(t,f) \odot |Y(t,f)|, \quad s=1...S
\end{equation}
where $Y$ is the frequency-domain representation of the mixed signal $y$, and $|Y|$ is computed through STFT transformation and represents the magnitude of the signal. $|\hat{X}_s|$ is also the magnitude of the $\hat{x}_s$ in the frequency-domain.
$ \hat{M}_s$ is the learned TF-mask function for speaker $s$ and $\odot$ is element-wise multiplication operation.
To reconstruct the signal in the time-domain, the estimated magnitude $|\hat{X}_s|$ and the phase of the mixed signal $\angle Y$ are given to ISTFT.
Throughout this paper all lower-case letters in equations represent signals in the time-domain and their upper-cases are representing the transformation into the frequency-domain.

\begin{figure}
\centering
\begin{subfigure}[b]{0.75\linewidth}
   \includegraphics[width=1\linewidth]{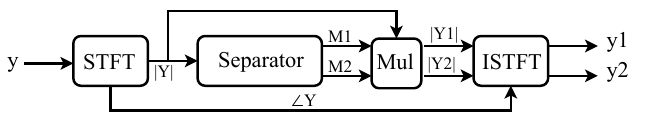}
   \caption{}
   \label{fig:specsep} 
\end{subfigure}
\begin{subfigure}[b]{0.75\linewidth}
   \includegraphics[width=1\linewidth]{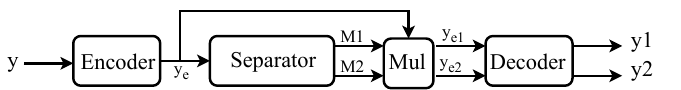}
   \caption{}
   \label{fig:wavesep}
\end{subfigure}
\caption[single-channel speech separation]{(a) Single-channel spectrogram speech separation (Mul is element-wise multiplication). (b) Single-channel waveform speech separation.}
\label{fig:sigsep}
\end{figure}

The block-diagram of single-channel frequency-domain speech separation is shown in Figure~\ref{fig:specsep}. Different separator networks \cite{chen2017deep} or loss functions \cite{erdogan2015phase, li2018cbldnn} and overall various methods \cite{wang2018supervised} are introduced before for single-channel frequency-domain speech separation, here we use \cite{kolbaek2017multitalker} as our baseline, which is using layers of BLSTM (bi-directional long short term memory) \cite{sak2014long} as the separator neural network along with uPIT-MSE (utterance level permutation invariant training-mean square error) loss function. The uPIT-MSE loss function is defined as,
\begin{equation}
\label{PIT}
J_{\phi^{*}} = \frac{1}{B} \sum_{s=1}^{S} \norm{ \hat{M}_s \odot \lvert Y \rvert - \lvert X_{\phi^{*}(s)} \rvert }_F^2.
\end{equation}
where ${\phi^{*}}$ is the permutation that minimizes the MSE error as,
\begin{equation}
\label{eq:min}
\phi^{*} = \argmin_{\phi \in \Phi } \frac{1}{B} \sum_{s=1}^{S} \norm{ \hat{M}_s \odot \lvert Y \rvert - \lvert X_{\phi(s)} \rvert}_F^2,
\end{equation}
and $\Phi$ is the set of all $S!$ permutations, $B=T \times F \times S$ is the total number of TF bins summed over all $S$ speakers, and $\norm{.}_F$ is the Frobenius norm.
This loss function can be improved with phase sensitive mean square error as in \cite{kolbaek2017multitalker}; however, for the sake of simplicity and emphasizing only on updating the magnitude component in frequency-domain solution, we use the Eq.~\ref{PIT} in our implementations.

The algorithm for single-channel speech separation is also summarized in Algorithm~\ref{algo:a} to facilitate following the materials covered in the next sections.

\begin{algorithm}
%\DontPrintSemicolon % Some LaTeX compilers require you to use \dontprintsemicolon    instead
\KwIn{y}
\KwOut{$x_s, \quad s=1:S$ (such as y = $\sum_{s=1}^{S}{x_s}$ )}
Calculate ($\lvert Y \rvert $, $\angle{Y}$) = STFT($y$) \\
Learn $\hat{M_s} = \textbf{F} (\lvert Y \rvert) \quad s=1:S$, with loss function in Eq.~\ref{PIT} (\textbf{F} refers to the separation model) \\
Iterate \textit{Step 2}, till current-epoch $<$ MAX-EPOCH \\
Calculate $\hat{x}_s$ using ISTFT($\hat{M_s} \odot \lvert Y \rvert$, $\angle{Y}$)
\caption{Single-Channel Speech Separation.}
\label{algo:a}
\end{algorithm}

\begin{figure}[t!]
  \centering
  \resizebox{7cm}{!}{
  \includegraphics[width=0.75\linewidth]{{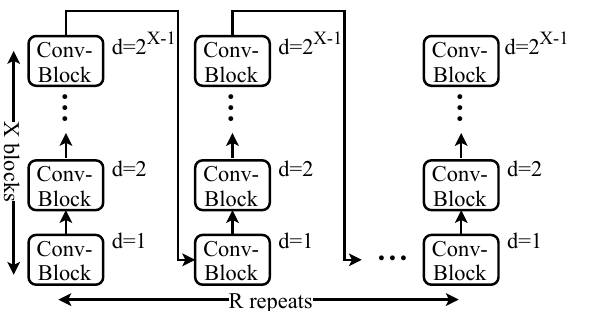}}}
  \caption{TCN-based separator network, where Conv-Block with increasing dilation (i.e., $d$) factor is cascade of [Conv-1D, PReLU \cite{he2015delving}, normalization layer, depth-wise convolution, PReLU, normalization layer] and it has skip connection from the input to the output. As input TCN receives the magnitude/encoded data and the TCN block follows by a convolution layer to output the masks.}
  \label{fig:TCN}
\end{figure}

\section{Unified Speech Separation Framework}
\label{sec:unified}

In this section, we present a unified framework for spectrogram and waveform separation with end-to-end training, which further is extended to multi-channel.

The block-diagram of a traditional spectrogram and waveform separation network is shown in Figure~\ref{fig:sigsep}. As our baseline and state-of-the-art solutions for spectrogram and waveform separation, we refer to \cite{kolbaek2017multitalker} and \cite{luo2019conv}, respectively. These two are mainly different in three aspects: 
(i) Encoder and decoder: 
in spectrogram separation STFT/ISTFT is applied to transform the data into/from the frequency-domain; while in waveform separation the encoding and decoding of the data is done with Conv-1D and ConvTranspose-1D operations which transform the data into/from a new temporal resolution. Therefore, the latter one does not decompose the data into magnitude and phase components, it is operating on a new temporal resolution. This can help to achieve better performance, the spectrogram separation only updates the magnitude component and phase is directly copied from the mixed signal.
(ii) The separator neural network: the baseline spectrogram and waveform separation networks used BLSTM and temporal convolutional network (TCN) \cite{lea2016temporal}, respectively. 
(iii) The loss function: the spectrogram separation is trained by uPIT-MSE loss function which minimizes the mean square error of the estimated magnitudes; and the waveform separation network directly optimizes the uPIT-SiSNR loss function.

\begin{figure*}[t!]
  \centering
  \resizebox{14cm}{!}{
  \includegraphics[width=0.75\linewidth]{{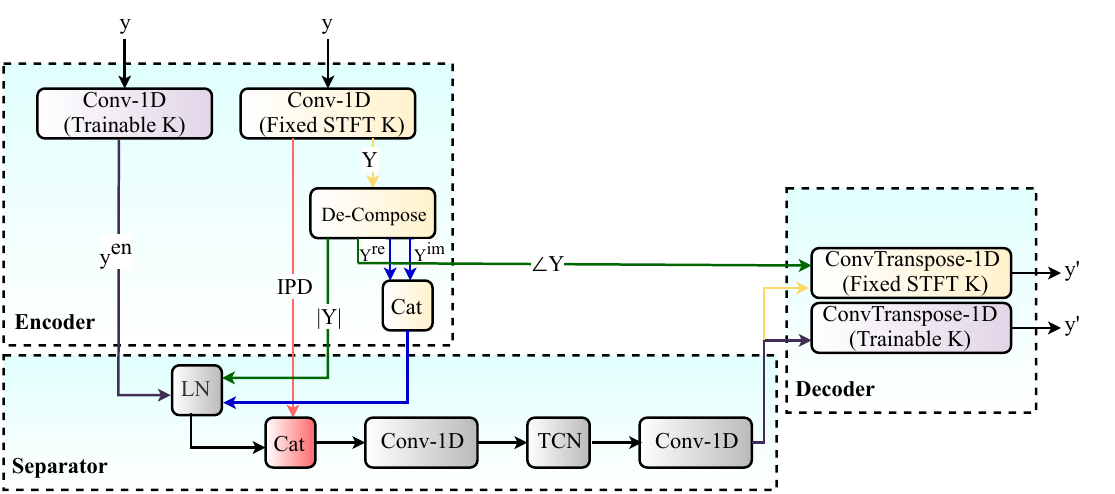}}}
  \caption{Unified multi-channel speech separation framework. LN is channel-wise layer normalization, and Cat represents the concatenation operation. Three different pipelines are shown with different colors, and the Separator is shared between all.}
  \label{fig:UNI}
\end{figure*}

In our proposed unified framework, we intend to integrate the uPIT-SiSNR loss function (as it directly optimizes the separation performance criterion) to learn the parameters, and it should be compatible with spectrogram separation setup as well. The traditional implementation of STFT and ISTFT limits the network to use this loss function. In Section~\ref{sec:e2e}, we propose a solution for updating the STFT/ISTFT implementation which simply enables the gradients to propagate backward in neural network through STFT and ISTFT layers as well.
\cite{luo2018tasnet} reported the proposed TCN separator network outperformed the one with BLSTM layers, for the speech separation task. Therefore, our unified framework uses TCN-based separator (however, it can be easily updated to another structure as well).
The block-diagram for TCN-based separator is shown in Figure~\ref{fig:TCN}.
In addition, we extend our unified structure to multi-channel data, adding spatial features as well which details are covered in Section~\ref{sec:multi}.
The overall block-diagram of the proposed unified multi-channel end-to-end speech separation is shown in Figure~\ref{fig:UNI}, the details are discussed in the following subsections.

\subsection{End-to-End Training}
\label{sec:e2e}

In this section, we re-formulate STFT and ISTFT transformations into a derivable format for the purpose of end-to-end training of the proposed unified framework.

For a time-domain signal $y$ (for simplicity, we assume $y$ is single-channel here; for multi-channel data STFT is applied separately on each individual channels), the STFT transformation generates complex-valued frequency-domain features $Y$ as follow:
\begin{equation}
\begin{split}
y[n] \xrightarrow[]{\tt STFT} {Y}_{nk}
&=\overset{T-1}{\underset{t=0}{\sum}}y[t]w[n-t]e^{-i\frac{2\pi t}{T}k}
\end{split}
    \label{eq:STFT}
\end{equation}
where $w$ is the window function with length $T$ (here we use Hann window), $n$ is the index of samples and $k$ is the index of frequency bands.
We can re-formulate the STFT operation in Eq.~\ref{eq:STFT} as,

\begin{equation} 
y[n] \xrightarrow[]{\tt STFT} {Y}_{nk}
=\overbrace{e^{-i\frac{2\pi n}{T}k}}^{\tt{phase~factor}} (y[n] \circledast \overbrace{w[n]e^{i\frac{2\pi n}{T}k}}^{\tt{STFT~kernel}} ),
    \label{eq:STFT_conv}
\end{equation}
where $\circledast$ is the convolution operation.

Based on Eq.~\ref{eq:STFT_conv} we can simply implement the STFT transformation throughout 1 layer of 1-D temporal convolution with an especial kernel function (as specified in the Eq.~\ref{eq:STFT_conv}) to match the traditional STFT computation.
At the reconstruction step, the ISTFT can be simply replaced with 1-D convolution transpose operation as well, using the same kernel function with ConvTranspose-1D layer (considering ISTFT is inverse of STFT operation). 

Therefore, the spectrogram encoder and decoder layers (STFT and ISTFT implementation with convolution) are different from the waveform separation encoder and decoder only in the kernel functions used in Conv-1D and ConvTrasnpose-1D. For the first one, kernel is fixed while for the latter one the kernel is trainable; in addition, the encoder and decoder do not share the same kernel function in waveform separation pipeline but for spectrogram they are tied together.

In the waveform separation, the output of the encoder is a real-valued vector, and is directly passed to the separation network.
For the spectrogram separation, the output of the encoder is a complex-valued vector, which is possible to be directly used as the input of the separator network (both real and imaginary parts can be given as the input to the network).  
Therefore, we expect complex-valued vector at the output of the decoder again, which is fine to directly fed them into the ConvTranspose-1D layer and reconstruct the signal.
In this scenario, the phase component as well is updating; therefore, at reconstruction it is unnecessary to pass the mixed signal phase.
However, to match the traditional spectrogram separation scenario (which de-composes the complex-valued frequency-domain representation of the signal into magnitude and phase), we can de-compose ${Y}_{nk}$ into magnitude $\lvert {Y}_{nk} \rvert$ and phase $\angle {Y}_{nk}$ components as well, and fed the magnitude to the separator. The magnitude and phase are calculated from the ${Y}_{nk}$ as follows ,

\begin{equation}
\label{eq:mg_ph}
\begin{aligned}
\lvert {Y}_{nk} \rvert &= \sqrt{{Y}^{\tt re}_{nk} + {Y}^{\tt im}_{nk}} , \\
\angle {Y}_{nk} &= \arctan ({Y}^{\tt im}_{nk}, {Y}^{\tt re}_{nk}).
\end{aligned}
\end{equation}

The $\lvert {Y}_{nk} \rvert$ is passed into the separator network to learn the mask functions and estimate the magnitude $\lvert \hat{X}_{nk} \rvert$ for source $x$. Next, ISTFT receives the estimated magnitude and phase of the mixed signal to reconstruct the results in the time-domain. The complex-valued signal is reconstructed as,

\begin{equation}
\label{eq:re_im}
\begin{aligned}
\hat{X}^{\tt re}_{nk} &= \lvert \hat{X}_{nk} \rvert \times \cos{\angle {Y}_{nk}} , \\
\hat{X}^{\tt im}_{nk} &= \lvert \hat{X}_{nk} \rvert \times \sin{\angle {Y}_{nk}}.
\end{aligned}
\end{equation}

As the network only separates the magnitude, the phase of the given mixed signal usually is employed to reconstruct the time-domain waveform.

Figure~\ref{fig:UNI} shows the unified structure for waveform and spectrogram separation. The gray boxes and black lines are shared between the two. The purple boxes and purple lines are only set active for the waveform separation. The yellow boxes, green and blue lines are for spectrogram separation. The blue lines can be set active to train the network on the complex-valued frequency-domain representation, or the green lines can be followed to have the same scenario as the traditional spectrogram separation.

\subsection{Multi-Channel Speech Separation}
\label{sec:multi}

In this section, we extend our unified structure for single-channel speech separation to an end-to-end multi-channel framework with including spatial features as well.

IPD (inter-microphone phase difference) features \cite{chen2018multi} represent cross-correlation between the channels and provide information about spatial location of the sources through calculating the time difference of arrival between the pre-defined microphone pairs. IPD features for the pair $u=<u_1, u_2>$ are calculated as,

\begin{equation}\label{eq:IPD}
\text{IPD}^{(u)}_{nk}=\angle{{Y}}^{u_1}_{nk}-\angle{{Y}}^{u_2}_{nk}, 
\end{equation}
where it computes the phase difference between microphone index $u_1$ and $u_2$. The IPDs are calculated on different microphone-pair sets for WSJ0 and LibriSpeech which details are provided in Section~\ref{sec:setup}.
In multi-channel experiments, all the IPD features (corresponding to all microphone-pairs) are concatenated to each other along with their $\cos{(.)}$ and $\sin{(.)}$.

The IPD features are computed using the phase components of the STFT analysis.
Therefore, to integrate the IPD calculation block into the end-to-end multi-channel training structure, we can first split the real and imaginary parts of the SFTF kernel in Eq.~\ref{eq:STFT_conv} as,

\begin{equation}
\begin{split}
{K}^{\tt re}_{nk} &=w[n] \cos(2\pi nk/T) \\
{K}^{\tt im}_{nk} &=w[n] \sin(2\pi nk/T)
\end{split}
\label{eq:STFTkernel}
\end{equation}
as mentioned earlier $w[n]$ is the window function used in the STFT analysis with length $T$.
Therefore, for the $u$-th pair, the IPD can be calculated as:

\begin{equation}
\text{IPD}^{(u)}_{nk}=\arctan\left( \frac{y^{u_1} \circledast {K}^{\tt re}_{nk}  }{y^{u_1} \circledast {K}^{\tt im}_{nk}  }\right)
-\arctan\left(
\frac{y^{u_2} \circledast {K}^{\tt re}_{nk}  }{y^{u_2} \circledast {K}^{\tt im}_{nk} }
\right).
\label{eq:kernel_IPD}
\end{equation}

Calculating the IPD with spectrogram separation framework is straightforward, as the phase of each channel is directly computed through the encoder along with the magnitude (i.e., the spectral features given as the input to the separator module). However, for the waveform separation different strategies can be applied: 
(1) Using the STFT analysis to compute the phase and magnitude and then calculate the IPD. However, there will be a mismatch between the IPD features and the encoder output and we may need to apply up-sampling on the IPD features in case the window-length/stride for calculating STFT and encoder convolution layers are different.
(2) Although the Eq.~\ref{eq:STFT_conv} and \ref{eq:STFTkernel} are derived from the STFT analysis; it can be generalized to any arbitrary kernel functions. It means that with any kernel function, we can still use Eq.~\ref{eq:kernel_IPD} to compute the IPD features.
Based on our initial experiments, we decided to choose the first strategy. The window-length/stride for STFT analysis as well is matched to the encoder window-length/stride, which requires no further down/up-sampling of the IPD features.
            
Figure~\ref{fig:UNI} shows the unified speech separation block-diagram including multi-channel extension highlighted with the red blocks and lines.

Here, we only use IPD features to extend the single-channel speech separation into multi-channel framework. There are other features as well that all can be integrated to benefit the task further, including speaker embeddings \cite{liu2019single} or angle features \cite{lorrytgt, fbinter} of each overlapping speaker towards a reference microphone for example. 
With the appended speaker-specific information, the speech separation task can be re-formulated to target speech extraction framework \cite{fbinter}. 
Target speech extraction only outputs the speech of the speaker of an interest; the information for the desired speaker can be provided through speaker location or speaker identity and it relaxes the prior knowledge on the exact number of overlapping speech sources \cite{lorrytgt}. 
The framework we proposed here can easily be extended to target speaker extraction as well.

\subsection{Spectrogram vs Waveform Separation}

In this section, we review our motivation for unifying the structure for spectrogram and waveform separation. In addition, we summarize the strengths and weaknesses of each setting based on our proposed framework.

Speech separation has been studied before usually on only frequency-domain representation or on a raw waveform. Here, we show these categorization is only different in the kernel function used for encoding the data, and can have a more flexible structure without limiting the task only to frequency-domain or time-domain. In either system configurations, there are some positive and some negative points which would be better to tune the representation-specific parameters as well based on the application and data. Our proposed framework is beyond these two representations and can be extended to complex-valued representations as well with any arbitrary kernel. This adds more hyper-parameters to the framework and provides a wider range of flexibilities. 

For example, to compare the frequency-domain and time-domain separation pipelines, we can numerate the following items that can either be counted as a strength or a weakness depending on the data:
\begin{itemize}
    \item The window-length and stride used for encoding the data. Our experiments show that longer window-length is better for frequency-domain and shorter for time-domain. With long window-size, data will be encoded into smaller number of frames which can be more efficient from the memory usage point of view. With shorter duration the latency will be shorter.
    \item The tied kernel function. In frequency-domain the kernel function used for encoder, decoder and spatial feature extractor are all tied together. However, in time-domain each have their own kernel functions, that introduces mismatch to the network setting. 
    \item Complex/real-valued encoded data. The frequency-domain encoder generates complex-valued output while for time-domain the output is in a new temporal resolution. Separating the complex-valued data for the purpose of updating the phase along with the magnitude can be more difficult than the time-domain. The latter one models the separation module on a unified representation without de-composing the input into magnitude and phase.
\end{itemize}

\begin{figure}
\centering
\begin{subfigure}[b]{0.55\linewidth}
   \includegraphics[width=1\linewidth]{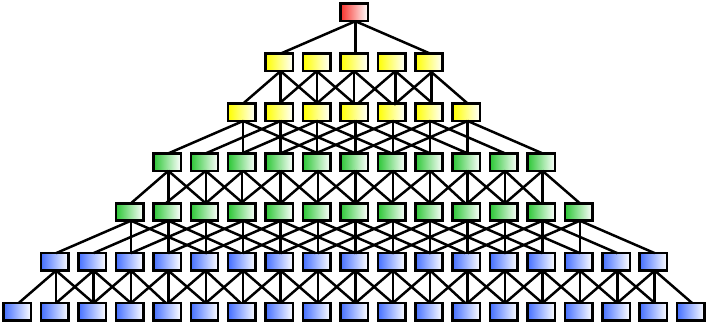}
   \caption{}
   \label{fig:non-causal} 
\end{subfigure}
\begin{subfigure}[b]{0.55\linewidth}
   \includegraphics[width=1\linewidth]{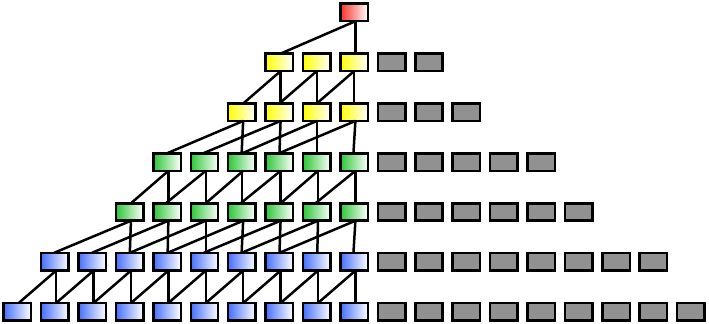}
   \caption{}
   \label{fig:causal}
\end{subfigure}
\begin{subfigure}[b]{0.55\linewidth}
   \includegraphics[width=1\linewidth]{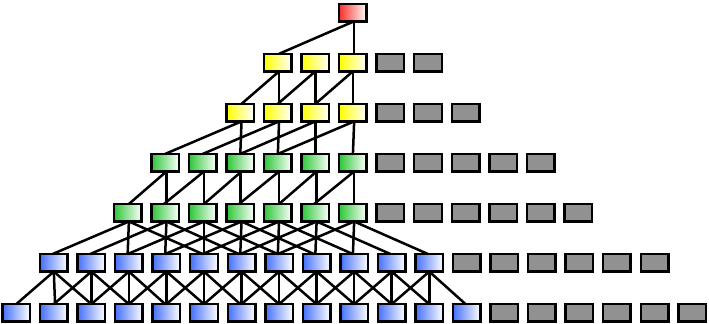}
   \caption{}
   \label{fig:semi-causal}
\end{subfigure}
\caption[]{(a) Non-causal, (b) Causal, (c) Semi-causal in dilated CNN.}
\label{fig:latency}
\end{figure}

\section{Latency}
\label{sec:latency}

The separation neural network shown in Figure~\ref{fig:TCN} has dilated convolution layers, where with deeper layers the dilation increases exponentially. The number of seen future frames is important in real applications where the user cannot tolerate a long latency.
One solution to control the latency problem can be using causal convolution over the current non-causal structure; \cite{luo2019conv} reported results for both causal and non-causal modes which showed the SDR performance degraded by almost 5dB (the SDR changes from 15.6 dB to 11 dB) when causal mode was used. It means that completely ignoring the future frames will result in significant loss in the SDR criterion for speech separation task.
Here, we introduce an asymmetric dilation structure to not only limit the latency but also preserve the performance as much as possible. The proposed solution is shown in Figure~\ref{fig:latency}.

Figure~\ref{fig:latency}, for simplicity shows a TCN separator network with hyper-parameters $R=3$, $X=2$ and $P=3$. Starting from the bottom layer the dilations are \{1, 2\}, \{1, 2\}, \{1, 2\}.
The Figure~\ref{fig:non-causal} shows the non-causal mode, the red frame at the top sees the whole future frames as defined by the dilation parameter. Figure~\ref{fig:causal} shows the same network but with causal mode, which in every frame no future frame has been seen.
As it is shown in \cite{luo2019conv} and we confirm in our experiments changing from non-causal to causal mode hurts the system performance significantly. 
In real-world applications (such as, including subtitles to movies), usually the user can tolerate a very small delay without even noticing. 
Therefore, we propose to limit the observation of future frames to only a few frames, and to have asymmetric dilation in the convolution layers as it is shown in figure~\ref{fig:semi-causal}. In the semi-causal mode (with asymmetric dilation), the non-causal mode is active only for both convolution blocks in $R=1$; however, in the deeper layers the causal mode has been set as True. Therefore, the latency is reduced by $R$ times.

\section{Experiments}
\label{sec:exp}

\begin{table*}[!th]
\scriptsize
\caption{{ \it Experimental setup for core developed models. }}
\centering
%\resizebox{14cm}{!}{
\begin{tabular}{c c c c c c c}
\hline
\textbf{Data} & \textbf{Domain} & \textbf{Model} & \textbf{Input} & \textbf{\# of Param.} &  \textbf{Setting} & \textbf{RF (s)} \\ \hline
WSJ0 & F & BLSTM & $|Y_0|$ & 67.05M & 4 $\times$ BLSTM-896 & - \\
WSJ0 & F & BLSTM & $|Y_0|$ + cos(IPD) + sin(IPD) & 89.15M & 4 $\times$ BLSTM-896 & - \\
WSJ0 & F & TCNN & $|Y_0|$ & 8.78M & L/N = 512/257 & 20.48 \\
WSJ0 & F & TCNN & $|Y_0|$ + cos(IPD) + sin(IPD) & 9.58M & L/N = 512/257$\times$13 & 20.48 \\
\hdashline
WSJ0 & T & TCNN & $y_0$ & 8.76M & L/N = 40/256 & 2.56 \\
WSJ0 & T & TCNN  & $y_0$ + cos(IPD) + sin(IPD) & 8.83M & L/N = 40/256$\times$13 & 2.56 \\
\hline
LS & F & TCNN & $|Y_0|$ & 16.88M & L/N = 512/257 & 122.88 \\
LS & F & TCNN & $|Y_0|$ + cos(IPD) + sin(IPD) & 17.80M & L/N = 512/257$\times$15 & 122.88 \\
\hdashline
LS & T & TCNN & $y_0$ & 16.27M & L/N = 80/257 & 30.7 \\
LS & T & TCNN  & $y_0$ + cos(IPD) + sin(IPD) & 16.41M & L/N = 80/257$\times$15 & 30.7 \\
\hline
\end{tabular}
%}
\label{tab:setup}
\vspace{-0.25cm}
\end{table*}

\subsection{Data}

We evaluate our speech separation models under different scenarios for both Wall Street Journal-0 (WSJ0) and LibriSpeech corpora, details on the simulation of these two sets of data are summarized in the next subsections.
The sampling rate for both datasets is 16KHz.

\subsubsection{WSJ0}
\label{sec:wsj}

We simulated a spatialized reverberant dataset derived from the Wall Street Journal 0 (WSJ0)-2mix corpus, which is a well-studied dataset in single and multi-channel speech separation. There are 20,000 (30 hr), 5,000 (8 hr) and 3,000 (5 hr) multi-channel, reverberant, two speaker mixed speech in training, validation and test subsets, respectively.
The utterance pairs chosen to create the mixed signals in the training, validation, and test sets are exactly based on \cite{hershey2016deep, luo2019conv}, only the simulation is different here; in \cite{hershey2016deep, luo2019conv} data was prepared for single-channel non-reverberate mixed signals.
The performance evaluation is all done on the test set, where all the speakers are unseen during training. We consider a 6-microphone circular array of 7-cm diameter, where speakers and the microphone-array are randomly located in the room. The two speakers and microphone array are on the same plane and all of them are at least 0.3m away from the wall. The image method \cite{allen1979image} is employed to simulate RIRs randomly from 3000 different room configurations with the size (length $\times$ width $\times$ height) ranging from 3m $\times$ 3m $\times$ 2.5m to 8m $\times$ 10m $\times$ 6m. The reverberation time T60 is sampled in a range of 0.05s to 0.5s. Samples with angle difference of 0-15\degree, 15-45\degree, 45-90\degree and 90-180\degree respectively account for 16\%, 29\%, 26\% and 29\% in the dataset.
From these 6-channels, we always use the first channel as input to the network representing the mixed signal.
For calculating the inter-microphone phase difference (IPD) as an input to our multi-channel models (described in detail in section~\ref{sec:multi}), we used the set of (1, 4), (2, 5), (3, 6), (1, 2), (3, 4), (5, 6) as the pairs of microphones. 
% ipd = "0,3;1,4;2,5;0,1;2,3;4,5"

\subsubsection{LibriSpeech}
\label{sec:ls}

To create reverberant multi-channel speaker mixtures from LibriSpeech corpus, two random utterances from two different speakers are convolved with the room impulse responses (RIRs), and then they are mixed together with different range of overlap to create the LibriSpeech-2mix dataset. 
We consider a 6-microphone array; speakers and the microphone array randomly located in the room. 
The two speakers and microphone array are at least 0.3m away from the wall; the distance between the speakers and the microphone array ranges between 0.5-6m. 
We employ image method \cite{allen1979image} to simulate RIRs randomly from 10000 different room configurations with the size (length $\times$ width $\times$ height) ranging from 3m $\times$ 3m $\times$ 2.5m to 10m $\times$ 8m $\times$ 6m. The reverberation time RT60 is sampled in a range of 0.05s to 0.7s. 
We generate 28,500 (125 hr), 5,000 (22 hr) and 3,000 (13 hr) 6-channel utterances for training, validation and testing, respectively. The test speakers are unseen in the training phase. The RIRs used in validation and testing are unseen in the training phase.
The first speaker is constrained to be from 225 degree to 315 degree (azimuth), i.e. maximum 45 degree away from 270 degree which is the front view of the microphone array.
No constraint has been considered on the interfering speaker or noise source in terms of azimuth.
%The distribution of the microphone array has been set within the room as [0, 0.045, 0; -0.012, 0.15, 0; -0.012, -0.015, 0; 0, -0.045, 0; 0.12, -0.03, 0; 0.12, 0.03, 0]. 
The microphone pairs utilized for computing IPDs (as described in detail in section~\ref{sec:multi}) are (1, 4), (2, 5), (3, 6), (2, 6), (3, 5), (1, 6), (4, 5). 
Samples with angle difference of 0-15\degree, 15-45\degree, 45-90\degree and 90-180\degree respectively account for 11\%, 20\%, 20\% and 49\% in the dataset.

\subsection{Metrics}

Different metrics are introduced for evaluation of speech separation systems \cite{vincent2006performance}. We use Si-SNR and SDR to objectively measure the separation accuracy.
\vspace{-2pt}
\begin{equation}
\label{sdr}
\text{SDR} = 10 \log_{10} \frac{\norm{x_{\sf target}}^2} { \norm{e_{\sf inter} + e_{\sf noise} + e_{\sf artif}}^2 },
\end{equation}

\begin{equation}
\label{si-snr}
\text{Si-SNR} = 10 \log_{10} \frac{\norm{x_{\sf target}}^2} { \norm{e_{\sf noise}}^2 },
\end{equation}
where $\sf  x_{target} = \frac{ <\hat{x} , x> x }{ ||x||^2 }$. $\hat{x}$ and $x$ respectively represent estimated and reference signals. For Si-SNR, the scale invariant is guaranteed by mean normalization of estimated and reference signals to zero mean \cite{luo2018tasnet}.

Quality of separated speech sources is also evaluated with PESQ (perceptual evaluation of speech quality) \cite{rix2001perceptual}.

\begin{table}[!th]
\scriptsize
\caption{{ \it Hyper-parameters for the unified framework.}}
\centering
\begin{tabular}{c c }
\hline
\textbf{Notation} & \textbf{Description} \\
\hline
N & Number of filters in encoder \\
L & Length of filter/window in encoder \\
B & Number of output channels in Conv-1D blocks in figure~\ref{fig:UNI} \\ 
R & Number of repeats in TCN (Figure~\ref{fig:TCN})\\
X & Number of Conv-Blocks in each R (Figure~\ref{fig:TCN})\\
P & Kernel size in Conv-Blocks (Figure~\ref{fig:TCN})\\
H & Number of channels in Conv-Blocks (Figure~\ref{fig:TCN})\\
\hline
\end{tabular}
\label{tab:hparam}
\vspace{-0.25cm}
\end{table}

\subsection{Experimental Setup}
\label{sec:setup}

Table \ref{tab:setup} summarizes the network setup and details on our core developed models (F- and T- mean frequency and time-domain in the tables). The receptive field (RF) seems very long for LS data, specially in F-domain which is mainly because of the stride-size of the encoder convolution layer.
The hyper-parameters for our unified model and separator (shown in Figure~\ref{fig:UNI} and Figure~\ref{fig:TCN}, respectively) are defined in Table~\ref{tab:hparam}.
For all the non-linear layers in our unified framework ReLU \cite{dahl2013improving} activation function is applied (unless otherwise stated), and Adam optimization \cite{kingma2014adam} is also selected for training the parameters.
%The learning rate starts from 0.0005 and if the performance does not improve on the validation set for 3 successive epochs, the learning rate is decreased by 50\% (??).
For the WSJ0 experiments the hyper-parameters for the TCN network is set as X/R/B/H/P=8/4/257/512/3, and for LibriSpeech (LS) as X/R/B/H/P=10/6/256/512/3. 
The chunk-size in WSJ0 experiments is 4s and in LibriSpeech experiments is 8s. This configuration has been set based on the best achieved performance. These are the default values, if in any experiment different configuration is studied, it is mentioned explicitly.
In following experiments, the SDR and Si-SNR scores are reported for different angle differences (0-15\degree, 15-45\degree, 45-90\degree and 90-180\degree, all) between the two speakers.

\subsection{Experimental Result}

In this section, experimental results are presented. Experiments are divided into different subsections to be easier to follow. % and make conclusions. At the end as well, the outcomes of our experiments are discussed.

\subsubsection{Ideal Masks}
\label{sec:im}
Table~\ref{tab:mask} includes Si-SNR, SDR and PESQ scores for four ideal masks representing the reference performance of speech separation on our simulated data. These ideal masks are IAM (ideal amplitude mask), IBM (ideal binary mask), IRM (ideal ratio mask), and IPSM (ideal phase sensitive mask) \cite{kolbaek2017multitalker}.
For LibriSpeech (LS), the ideal mask scores are better comparing against the WSJ0, which is related to the simulation configuration. In LibriSpeech, a larger percentage of data are allocated to the larger angle differences between the two speakers.

\begin{table}[!th]
\scriptsize
\caption{{ \it Ideal mask scores on WSJ0 and LS.}}
\centering
\begin{tabular}{c c c c c}
\hline
\textbf{Data} & \textbf{Ideal Mask} & \textbf{Si-SNR} & \textbf{SDR} & \textbf{PESQ} \\
\hline
\multirow{4}{*}{WSJ0 / LS} & IAM & 11.04 / 13.07 & 11.3 / 13.08 & 3.78 / 3.81\\
& IBM  & 11.53 / 12.87 & 11.88 / 12.94 & 2.46 / 2.58 \\
& IRM  & 10.98 / 12.39 & 11.4 / 12.49  & 3.62 / 3.49 \\
& IPSM & 13.63 / 15.85 & 13.98 / 15.87 & 3.72 / 3.75 \\
\hline
\end{tabular}
\label{tab:mask}
\vspace{-0.25cm}
\end{table}

\subsubsection{Baseline}

The models introduced in \cite{kolbaek2017multitalker} and \cite{luo2019conv} (i.e., uPIT-MSE for frequency-domain with BLSTM network, and Conv-TasNet with TCN network for time-domain) are chosen here as the baseline for spectrogram and waveform separation, respectively. 
Table~\ref{tab:baseline} shows the baseline solution results on our simulated data.
Most of previous research studies are limited to single-channel WSJ0-2mix where no reverberation is considered for simulating the data; therefore, we cannot directly compare their performance with our results here. We re-implemented those solutions and results are presented here.
The time-domain baseline uses gLN (global layer normalization), as it has been reported before in \cite{luo2019conv} that outperforms other normalization techniques. Here, we set the hyper-parameters to achieve the best comparable performance on our simulated data.
As we expected, the time-domain speech separation performs better than frequency-domain. We believe, this experiment is not fair enough to conclude the superiority of the time-domain separation over the frequency-domain. In continue, we extend our experiments to multi-channel framework. In addition, the effective components of Conv-TasNet are integrated into the frequency-domain solution. 
It is worth mentioning that baseline frequency-domain solution uses uPIT-MSE (mean square error) as the objective function while training the separator, which directly does not optimize the separation performance criterion.

\begin{table*}[!th]
\scriptsize
\caption{{ \it Baseline spectrogram and waveform single-channel speech separation performance.}}
\centering
%\resizebox{14cm}{!}{
\begin{tabular}{c c c c  c  c c c c c c}
\hline
\textbf{Data} & \textbf{Domain} & \textbf{Model} & \textbf{Input} & \textbf{\# of Param.} &  \textbf{Setting} & \textbf{Norm.} & \textbf{RF (s)} & \textbf{Si-SNR (dB)} & \textbf{SDR (dB)} & \textbf{PESQ} \\ 
\hline
WSJ0 & F & BLSTM & $|Y_0|$ & 67.05M & 4 $\times$ BLSTM-896 & - & - & 6.71 & 7.16 & 2.43 \\
\hdashline
WSJ0 & T & TCN & $y_0$ & 8.76M & L/N=40/256 & gLN & 2.56 & 9.47 & 9.97 & 2.75\\
\hline
LS & F & BLSTM & $|Y_0|$ & 67.05M & 4 $\times$ BLSTM-896 & - & - & 4.12 & 4.46 &  2.04 \\ % ls_blstm_mse
\hdashline
LS & T & TCN & $y_0$ & 16.27M & L/N=80/256 & gLN & 30.7 & 7.97 & 8.40 & 2.63 \\ % tasnet_1ch_ls_g_gln
\hline
\end{tabular}
%}
\label{tab:baseline}
\vspace{-0.25cm}
\end{table*}

\subsubsection{Improved Spectrogram Separation}

We performed preliminary experiments on WSJ0 data with incorporating the TCN separation network (Figure~\ref{fig:TCN}) and uPIT-SiSNR loss into frequency-domain speech separation, for both single-channel (1-ch) and multi-channel (m-ch) frameworks.
The results are presented in Table~\ref{tab:init}.
The results confirm that TCN separation network and uPIT-SiSNR loss function improve the performance on spectrogram separation, consistently. The multi-channel models are also shown to outperform the single-channel models, for all the settings.
PESQ scores are not as consistent as the SDR and Si-SNR scores; however, for multi-channel models the PESQ scores are better, as expected.
In addition, for different angle differences, the scores do not change a lot with single-channel models; but in multi-channel models for larger angle differences the scores are better.
For smaller angles the two speakers are sitting more closer to each other, and separating their speech is more difficult when the IPD features are not accurate enough. 

In the following experiments we extend our evaluations to LibriSpeech data as well, and study the effectiveness of different hyper-parameters for the time- and frequency-domain solutions.
From now on, all our experiments are just limited to the TCN separation network with uPIT-SiSNR loss function; as it is shown in the unified framework.

\begin{table*}[!th]
\scriptsize
\caption{{ \it Improved spectrogram speech separation on WSJ0 with incorporating TCN separator and uPIT-SiSNR loss. }}
\centering
\begin{tabular}{c c c | c c c c | c | c c c c | c | c}
\hline
\multirow{2}{*}{\textbf{\# Ch}} & \multirow{2}{*}{\textbf{Model}} & \multirow{2}{*}{\textbf{Loss}} & \multicolumn{5}{c|}{\textbf{Si-SNR}} & \multicolumn{5}{c}{\textbf{SDR}} & \multirow{2}{*}{\textbf{PESQ}}\\
&  &  & 0-15\degree & 15-45\degree & 45-90\degree & 90-180\degree & \textbf{AVG} & 0-15\degree & 15-45\degree & 45-90\degree & 90-180\degree & \textbf{AVG} &  \\
\hline
\multirow{4}{*}{1-ch} & \multirow{2}{*}{F-BLSTM-1} & uPIT-SiSNR & 7.54 & 7.80 & 7.72 & 7.81 & \textbf{7.74} & 8.14 & 8.39 & 8.29 & 8.38 & \textbf{8.32} & 2.30 \\
 & & uPIT-MSE & 6.50 & 6.79 & 6.67 & 6.78 & \textbf{6.71} & 6.98 & 7.24 & 7.11 & 7.22 & \textbf{7.16} & 2.43 \\
& \multirow{2}{*}{F-CNN-1} & uPIT-SiSNR & 7.08 & 7.48 & 7.45 & 7.48 & \textbf{7.42} & 7.70 & 8.06 & 8.02 & 8.06 & \textbf{8.00} & 2.32 \\
& & uPIT-MSE & 6.31 & 6.67 & 6.54 & 6.65 & \textbf{6.58} & 6.81 & 7.13 & 7.0 & 7.1 & \textbf{7.04} & 2.48 \\
\hline
\multirow{4}{*}{m-ch} & \multirow{2}{*}{F-BLSTM-2} & uPIT-SiSNR & 5.41 & 9.37 & 10.13 & 10.65 & \textbf{9.38} & 6.13 & 9.89 & 10.62 & 11.13 & \textbf{9.91}  & 2.44 \\
& & uPIT-MSE &  5.68 & 8.50 & 9.09 & 9.17 & \textbf{8.45} & 6.17 & 8.88 & 9.45 & 9.53 & \textbf{8.83} & 2.73 \\
& \multirow{2}{*}{F-CNN-2} & uPIT-SiSNR &  6.88 & 10.27 & 11.02 & 11.54 & \textbf{10.36} & 7.5 & 10.75 & 11.47 & 11.99 & \textbf{10.84} & 2.56 \\
& & uPIT-MSE & 4.86 & 8.71 & 9.43 & 10.02 & \textbf{8.74} & 5.38 & 9.09 & 9.79 & 10.37 & \textbf{9.12} & 2.78 \\
\hline
\end{tabular}
\label{tab:init}
\vspace{-0.25cm}
\end{table*}

\subsubsection{Single-Channel Speech Separation}

Single-channel results for both WSJ0 and LibriSpeech (LS) are reported in Table~\ref{tab:1ch}.
For WSJ0, the spectrogram separation (denoted as F-domain in the table) achieves worst results comparing against the T-domain (i.e., time-domain).

Different configurations are tested on LS. Based on the results, we can observe that, (1) the BN (batch normalization) \cite{ioffe2015batch} gains better performance comparing against the gLN (global layer normalization) in contrast to the observation reported in \cite{luo2019conv}. (2) The F-domain results are better than T-domain, in contrast to WSJ0 scores. (3) As parameter $L$ in F-domain can be set to a larger value than T-domain, the chunk-size can also be set to a larger length for F-domain without requiring a lot more resources, and results show that with larger chunk-size the F-domain can gain better performance. 
We should mention that, the STFT encoder requires the $L$ parameter to be large enough to perform the analysis while the T-domain with trainable kernel performs better with $L$ set to a smaller value. In our experiments $L=512$ for F-domain and $L=80$ for T-domain is used. 

From one other view point, the training data for LS is almost 4 times larger than the WSJ0; we may conclude that with more training data, the F-domain can achieve better performance as well. It all depends on the resources and data available to choose between the two settings (i.e., spectrogram or waveform separation), and they both can have comparable results in terms of the speech separation criteria.
The PESQ scores here are not always consistent with the SDR and Si-SNR, but their trend is confirming the general conclusion made on the table.  

For the following experiments, we use the BN unless otherwise stated.

\begin{table*}[!th]
\scriptsize
\caption{{ \it Single-channel (1-ch) speech separation under various configurations for WSJ0 and LS.}}
\centering
\begin{tabular}{c c c c | c c c c | c | c c c c | c | c }
\hline
\multirow{2}{*}{\textbf{Data}} & \multirow{2}{*}{\textbf{Domain}} & \multirow{2}{*}{\textbf{CSZ}} & \multirow{2}{*}{\textbf{Norm.}} & \multicolumn{5}{c|}{\textbf{Si-SNR}} & \multicolumn{5}{c|}{\textbf{SDR}} & \multirow{2}{*}{\textbf{PESQ}} \\
& & & & 0-15\degree & 15-45\degree & 45-90\degree & 90-180\degree & \textbf{AVG} & 0-15\degree & 15-45\degree & 45-90\degree & 90-180\degree & \textbf{AVG} &  \\
\hline
\multirow{2}{*}{WSJ0} & T & 4s & gLN & 9.02 & 9.33 & 9.59 & 9.71 & \textbf{9.47} & 9.57 & 9.83 & 10.09 & 10.2 & \textbf{9.97} & 2.75 \\
& F & 4s & gLN & 7.08 & 7.48 & 7.45 & 7.48 & \textbf{7.42} & 7.70 & 8.06 & 8.02 & 8.06 & \textbf{8} & 2.32 \\
\hline
\multirow{6}{*}{LS} & T & 8s  & BN & 8.07 & 8.62 & 8.87 & 8.84 & \textbf{8.72} & 8.42 & 8.99 & 9.24 & 9.18 & \textbf{9.08} & 2.69 \\ % tasnet_1ch_ls/g
& T & 8s & gLN & 7.52 & 7.99 & 8.07 & 8.03 & \textbf{7.97} & 7.95 & 8.43 & 8.51 & 8.44 & \textbf{8.40} & 2.63 \\ % tasnet_1ch_ls/g_gln
& F & 8s & BN & 8.07 & 8.75 & 8.84 & 8.85 & \textbf{8.74} & - 8.47 & 9.17 & 9.26 & 9.26 & \textbf{9.16} & 2.56 \\  % freq_ls_mch_trainable_a_2_1ch
& F & 20s & BN & 8.93 & 9.52 & 9.46 & 9.42 & \textbf{9.40} & 9.42 & 9.99 & 9.96 & 9.88 & \textbf{9.87} & 2.59 \\ % freq_ls_mch_trainable_a_2_1ch_20s
& F & 8s & gLN & 8.34 & 8.84 & 8.97 & 8.87 & \textbf{8.83} & 8.76 & 9.30 & 9.43 & 9.31 & \textbf{9.27} & 2.60 \\ % freq_ls_mch_trainable/a_2_1ch_gln
& F & 20s & gLN & 7.86 & 8.36 & 8.31 & 8.31 & \textbf{8.27} & 8.34 & 8.83 & 8.80 & 8.75 & \textbf{8.73} & 2.48 \\ % freq_ls_mch_trainable/a_2_1ch_20s_gln
\hline
\end{tabular}
\label{tab:1ch}
\vspace{-0.25cm}
\end{table*}

\subsubsection{Multi-Channel Speech Separation}

Table~\ref{tab:mch} compares the multi-channel results in T- and F-domain on both WSJ0 and LS. For WSJ0, the waveform separation obtains better performance; however, for LS, spectrogram separation performs slightly better.
Comparing Table~\ref{tab:1ch} and Table~\ref{tab:mch}, obviously confirms the superiority of multi-channel framework over the single-channel framework, and the effectiveness of the proposed multi-channel structure.
The following experiments are all just evaluated on the LS data with multi-channel framework. The LS data is larger than WSJ0 and developed systems are more robust which makes the conclusions more reliable and concrete.

\begin{table*}[!th]
\scriptsize
\caption{{ \it Multi-channel (m-ch) speech separation for WSJ0 and LS.}}
\centering
\begin{tabular}{c c | c c c c | c | c c c c | c | c }
\hline
\multirow{2}{*}{\textbf{Data}} & \multirow{2}{*}{\textbf{Domain}} & \multicolumn{5}{c|}{\textbf{Si-SNR}} & \multicolumn{5}{c|}{\textbf{SDR}} & \multirow{2}{*}{\textbf{PESQ}} \\
& & 0-15\degree & 15-45\degree & 45-90\degree & 90-180\degree & \textbf{AVG} & 0-15\degree & 15-45\degree & 45-90\degree & 90-180\degree & \textbf{AVG} &  \\
\hline
\multirow{2}{*}{WSJ0} & T & 7.70 & 11.63 & 12.33 & 12.62 & \textbf{11.55} & 8.31 & 12.07 & 12.74 & 13.03 & \textbf{11.99} & 2.97 \\
& F & 6.88 & 10.27 & 11.02 & 11.54 & \textbf{10.36} & 7.5 & 10.75 & 11.47 & 11.99 & \textbf{10.84} & 2.56 \\
\hline
\multirow{2}{*}{LS} & T & 8.08 & 10.59 & 11.57 & 12.68 & \textbf{11.55} & 8.38 & 10.88 & 11.84 & 12.92 & \textbf{11.82} & 3.02 \\ % tasnet_mch_ls_9
& F & 8.34 & 10.69 & 11.52 & 12.64 & \textbf{11.57} & 8.70 & 11.02 & 11.90 & 12.93 & \textbf{11.89} & 2.76 \\ %  a_2
\hline
\end{tabular}
\label{tab:mch}
\vspace{-0.25cm}
\end{table*}

Table~\ref{tab:mextend} extends the multi-channel experiments on LS for different system configurations. Different values for $L$ and chunk-size (CSZ) are experimented here.
In F-domain setup the CSZ can be easily changed to a larger value without requiring additional resources. 
The window-length and stride is almost long for spectrogram separation pipeline; therefore, encoded data includes less number of frames comparing against the waveform separation. This makes the spectrogram separation more efficient in terms of memory-usage.
The T-domain scores for $L=512$ is also reported here to confirm that performance degrades a lot when the $L$ is set to a larger value. Therefore, when there is memory limitations, F-domain setup would be preferred. In the T-domain, when the $L=512$, the larger chunk-size performs better; as the presented context is larger. However, the network setup limits the receptive field; therefore, we can not expect to gain better results with making the chunk-size as large as possible. In other words, when the chunk-size is within the receptive field, the network can gain benefits with longer chunk-size.

\begin{table*}[!th]
\scriptsize
\caption{{ \it Multi-channel (m-ch) speech separation with different settings on LS.}}
\centering
\begin{tabular}{c c c| c c c c | c | c c c c | c | c }
\hline
\multirow{2}{*}{\textbf{Domain}} & \multirow{2}{*}{\textbf{L}} & \multirow{2}{*}{\textbf{CSZ}} & \multicolumn{5}{c|}{\textbf{Si-SNR}} & \multicolumn{5}{c|}{\textbf{SDR}} & \multirow{2}{*}{\textbf{PESQ}} \\
& & & 0-15\degree & 15-45\degree & 45-90\degree & 90-180\degree & \textbf{AVG} & 0-15\degree & 15-45\degree & 45-90\degree & 90-180\degree & \textbf{AVG} &  \\
\hline
T & 80  & 8s & 8.08 & 10.59 & 11.57 & 12.68 & \textbf{11.55} & 8.38 & 10.88 & 11.84 & 12.92 & \textbf{11.82} & 3.02 \\ % tasnet_mch_ls_9
T & 80  & 12s & 5.80 & 10.13 & 11.25 & 12.57 & \textbf{11.09} & 6.25 & 10.47 & 11.56 & 12.84 & \textbf{11.41} & 2.97 \\ % 9_12s
T & 512 & 8s & 4.45 & 7.10 & 9.04 & 9.96 & \textbf{8.62} & 5.14 & 7.74 & 9.77 & 10.66 & \textbf{9.31} & 2.70 \\ % tasnet_mch_ls_9_512L
T & 512 & 12s & 7.67 & 9.58 & 9.92 & 10.57 & \textbf{9.94} & 8.25 & 10.14 & 10.51 & 11.14 & \textbf{10.51} & 2.79 \\ % tasnet_mch_ls_9_512L_12s
\hline
F & 512 & 8s & 8.34 & 10.69 & 11.52 & 12.64 & \textbf{11.57} & 8.70 & 11.02 & 11.90 & 12.93 & \textbf{11.89} & 2.76 \\ % a_2
F & 512 & 12s & 9.72 & 11.79 & 12.14 & 12.96 & \textbf{12.22} & 10.11 & 12.14 & 12.52 & 13.26 & \textbf{12.55} & 2.76 \\ % a_2_12s
F & 512 & 20s & 9.23 & 11.64 & 11.98 & 12.86 & \textbf{12.05} & 9.72 & 12.03 & 12.39 & 13.21 & \textbf{12.44} & 2.74 \\ % a_2_20s
\hline
\end{tabular}
\label{tab:mextend}
\vspace{-0.25cm}
\end{table*}

\subsubsection{Complex Input \& Trainable Window for Spectrogram Speech Separation}

All the experiments on spectrogram separation we reported here are based on the green pipeline shown in Figure~\ref{fig:UNI}. It means that, the magnitude component is given as the input to the TCN separator. And at the decoder, the phase of the mixed signal is employed to reconstruct the signal in time-domain. 
The other way to perform the spectrogram separation is to extract real and imaginary parts of the signal through STFT and use the concatenation of both to the separation network, i.e., the blue pipeline presented in Figure~\ref{fig:UNI}.
In this subsection, the scores achieved with complex-input to the speech separation network is reported. 
In addition, to add more flexibility to the spectrogram separation framework, we evaluated the system performance when the window function in calculating the kernel is trainable (instead of the fixed Hann window).
Table~\ref{tab:complex} reports the scores with these two settings added to the network.
Results are very competitive against the reference system for trainable window. 

Although we expected to gain more improvement with complex-valued input, we can justify the scores by (1) using the complex-valued data to train the current unified framework can be a difficult task, and may require more constraints to be included while training the parameters. (2) The hyper-parameters need to be adjusted (now, we evaluated the model based on the hyper-parameters set for the real-valued data). (3) The activation function used at the mask layer is not be proper for the new feature representation. 
This set of experiments require more investigation in future studies.

\begin{table*}[!th]
\scriptsize
\caption{{ \it Complex vs only magnitude input to TCN and trainable window for LS on m-ch framework.}}
\centering
\begin{tabular}{c c c| c c c c | c | c c c c | c }
\hline
\multirow{2}{*}{\textbf{Domain}} & \multirow{2}{*}{\textbf{Window}} & \multirow{2}{*}{\textbf{Input}} & \multicolumn{5}{c|}{\textbf{Si-SNR}} & \multicolumn{5}{c}{\textbf{SDR}} \\ % & \multirow{2}{*}{\textbf{PESQ}} \\
& & & 0-15\degree & 15-45\degree & 45-90\degree & 90-180\degree & \textbf{AVG} & 0-15\degree & 15-45\degree & 45-90\degree & 90-180\degree & \textbf{AVG} \\ % &  \\
\hline
F & Fixed & $|Y_0|$ + cos(IPD) + sin(IPD) & 8.34 & 10.69 & 11.52 & 12.64 & \textbf{11.57} & 8.70 & 11.02 & 11.90 & 12.93 & \textbf{11.89} \\ % & 2.76 \\ %  a_2
F & Fixed & $Y_0$ + cos(IPD) + sin(IPD) & 7.50 & 8.94 & 9.36 & 9.17 & \textbf{8.98} & 7.94 & 9.34 & 9.77 & 9.55 & \textbf{9.38} \\ % & - \\ % a_2_cc--
F & Trainable & $|Y_0|$ + cos(IPD) + sin(IPD) & 7.81 & 10.76 & 11.35 & 12.43 & \textbf{11.39} & 8.23 & 11.11 & 11.72 & 12.72 & \textbf{11.72} \\ % & 2.76 \\ % a_2_t3
\hline
\end{tabular}
\label{tab:complex}
\vspace{-0.25cm}
\end{table*}

\subsubsection{Latency}

Two different strategies can be considered to control the latency: (1) restricting the receptive field of the network, (2) using the causal mode. In addition, we proposed the third solution to use asymmetric dilation, that just limits the dilation over the future unseen frames.

Table~\ref{tab:latency1} shows the results for the first solution. We performed experiments on different network settings with changing the hyper-parameters $X$ and $R$ specifically, which directly are defining the receptive field of the TCN network.
In T-domain, we observe a degradation in the performance when a smaller network is trained. It shows that waveform separation performance is very dependent to the receptive field of the network, and scores degrade significantly with smaller networks.
On the other hand, the spectrogram separation framework is shown to be more stable over different network structures.
The table interestingly shows that F-domain solution in contrast to T-domain is more stable; specially for the smaller angle differences between the two speakers. 

Scores on second and third solutions for controlling the latency are reported in Table~\ref{tab:latency2}. 
As we expected, setting the causal mode is hurting the system performance significantly as it was observed in \cite{luo2019conv} as well. However, our proposed solution to have asymmetric dilation achieves a very competitive performance against the non-causal mode and it reduces the latency by 10 times (or generally speaking by $R$ times).

\begin{table*}[!th]
\scriptsize
\caption{{ \it Different network structures to limit the receptive-field for controlling the latency. Results are for m-ch framework on LS. The chunk-size is 8s and for T-domain L=80 and for F-domain L=512. }}
\centering
\begin{tabular}{c c c c | c c c c | c | c c c c | c | c }
\hline
\multirow{2}{*}{\textbf{Domain}} & \multirow{2}{*}{\textbf{RF (s)}} & \multirow{2}{*}{\textbf{X}} & \multirow{2}{*}{\textbf{R}} & \multicolumn{5}{c|}{\textbf{Si-SNR}} & \multicolumn{5}{c|}{\textbf{SDR}} & \multirow{2}{*}{\textbf{PESQ}} \\
& & & & 0-15\degree & 15-45\degree & 45-90\degree & 90-180\degree & \textbf{AVG} & 0-15\degree & 15-45\degree & 45-90\degree & 90-180\degree & \textbf{AVG} &  \\
\hline
T & 5.12 & 8 & 4 & 3.72 & 7.46 & 9.43 & 11.37 & \textbf{9.39} & 4.14 & 7.79 & 9.73 & 11.66 & \textbf{9.70} & 2.78 \\ % 9_X8R4
T & 12.8 & 8 & 10 & 3.15 & 7.44 & 10.42 & 12.19 & \textbf{9.92} & 3.72 & 7.83 & 10.72 & 12.46 & \textbf{10.26} & 2.80 \\ % 9_X8R10
T & 30.7 & 10 & 6 & 8.08 & 10.59 & 11.57 & 12.68 & \textbf{11.55} & 8.38 & 10.88 & 11.84 & 12.92 & \textbf{11.82} & 3.02 \\ %  tasnet_mch_ls_9
\hline
F & 20.48 & 8 & 4 & 8.35 & 10.78 & 11.64 & 12.66 & \textbf{11.62} & 8.75 & 11.10 & 11.99 & 12.94 & \textbf{11.94} & 2.77 \\ % a_2_X8X4
F & 51.2 & 8 & 10 & 9.13 & 11.51 & 11.91 & 12.78 & \textbf{11.96} & 9.52 & 11.82 & 12.27 & 13.07 & \textbf{12.28} & 2.81 \\ % a_2_X8X10
F & 122.88 & 10 & 6 & 8.34 & 10.69 & 11.52 & 12.64 & \textbf{11.57} & 8.70 & 11.02 & 11.90 & 12.93 & \textbf{11.89} & 2.76 \\ % a_2
\hline
\end{tabular}
\label{tab:latency1}
\vspace{-0.25cm}
\end{table*}

\begin{table*}[!th]
\scriptsize
\caption{{ \it Multi-channel speech separation on LS for causal, non-causal and semi-causal modes.}}
\centering
\begin{tabular}{c c | c c c c | c | c c c c | c | c }
\hline
\multirow{2}{*}{\textbf{Domain}} & \multirow{2}{*}{\textbf{Causal}} & \multicolumn{5}{c|}{\textbf{Si-SNR}} & \multicolumn{5}{c|}{\textbf{SDR}} & \multirow{2}{*}{\textbf{PESQ}} \\
& & 0-15\degree & 15-45\degree & 45-90\degree & 90-180\degree & \textbf{AVG} & 0-15\degree & 15-45\degree & 45-90\degree & 90-180\degree & \textbf{AVG} &  \\
\hline
T & False & 8.08 & 10.59 & 11.57 & 12.68 & \textbf{11.55} & 8.38 & 10.88 & 11.84 & 12.92 & \textbf{11.82} & 3.02 \\ % tasnet_mch_ls_9
T & True & 2.39 & 6.01 & 8.73 & 10.14 & \textbf{8.21} & 2.90 & 6.40 & 9.05 & 10.44 & \textbf{8.55} & 2.63 \\ % 9_causal
T & Semi & 8.18 & 11.13 & 11.74 & 12.78 & \textbf{11.75} & 8.46 & 11.36 & 11.99 & 12.99 & \textbf{11.98} & 2.99 \\ % 9_semicausal
\hline
F & False & 8.34 & 10.69 & 11.52 & 12.64 & \textbf{11.57} & 8.70 & 11.02 & 11.90 & 12.93 & \textbf{11.89} & 2.76 \\ %  a_2
F & True & 4.37 & 7.81 & 10.02 & 11.51 & \textbf{9.71} & 4.83 & 8.13 & 10.29 & 11.76 & \textbf{10.00} & 2.56 \\ % a_2_causal
F & Semi & 7.84 & 10.51 & 11.66 & 12.70 & \textbf{11.54} & 8.24 & 10.87 & 12.04 & 13.01 & \textbf{11.88} & 2.73 \\ % a_2_partialcausal
\hline
\end{tabular}
\label{tab:latency2}
\vspace{-0.25cm}
\end{table*}

\subsection{Discussion}

\begin{figure}[t!]
  \centering
  \resizebox{7cm}{!}{
  \includegraphics[width=0.75\linewidth]{{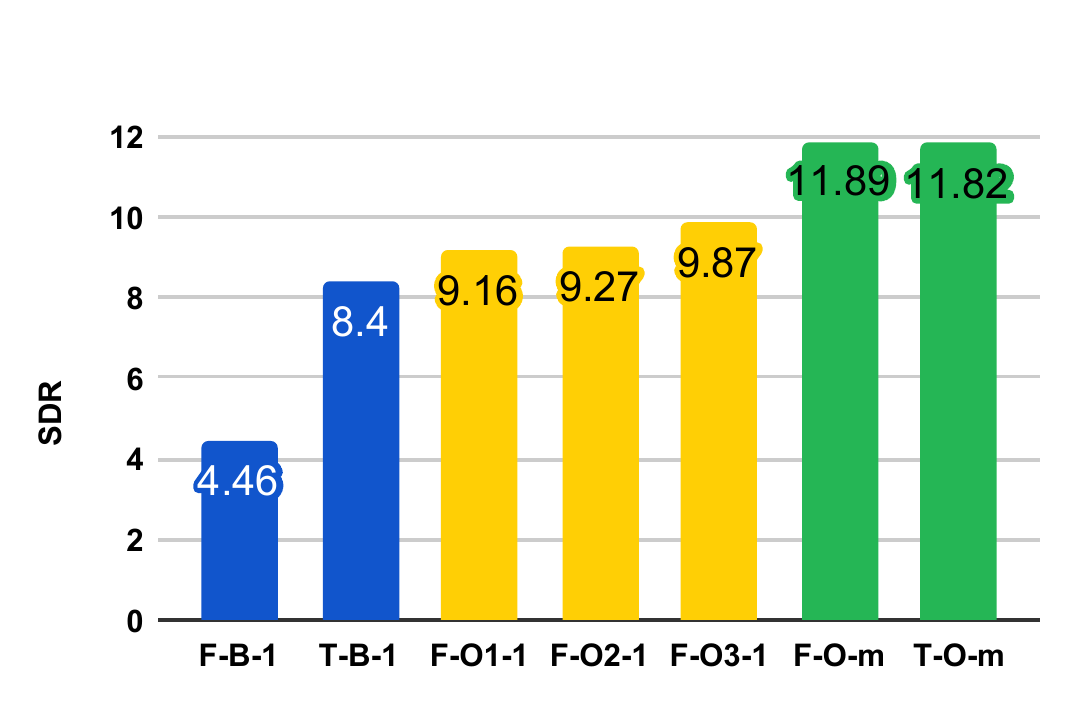}}}
  \caption{Improvements over the baseline (the name for the models are in the format: domain-baseline/ours-\#ch). The blue columns are for the baselines both time and frequency domain, the yellow columns are for our developed frequency-domain solutions for single-channel framework. The green colors are for multi-channel, where F-O-m is for frequency-domain and T-O-m is for time-domain.}
  \label{fig:chart1}
\end{figure}

As mentioned earlier, no other previous studies on speech separation had reported results for multi-channel reverberate data, on LibriSpeech corpus. Therefore, we re-implemented two state-of-the-art solutions in frequency-domain \cite{kolbaek2017multitalker} and time-domain \cite{luo2019conv} on our simulated data.
Figure~\ref{fig:chart1} represents the SDR scores and gains achieved with our proposed solutions for single-channel spectrogram speech separation; in addition, the SDR scores obtained with multi-channel frameworks are included in both time and frequency-domain.
The chart shows that using effective components of Conv-TasNet \cite{luo2019conv} can improve the speech separation performance in frequency-domain as well. With further tuning the hyper-parameters can improve the performance more on single-channel frequency-domain solutions.
Multi-channel spectrogram and waveform speech separation as well are shown to outperform the baseline systems.

In all the evaluated experiments, we also reported the PESQ scores. Although, the scores are not as consistent as the other reported criteria, the trend on PESQ scores matches the overall observation on speech separation accuracy.

In this study, our motivation was presenting a unified structure for speech separation, combining single/multi-channel with time/frequency-domain input representations. 
In addition, we designed experiments and simulated diverse datasets to comprehensively evaluate each pipeline through the unified framework. The results show that even for single-channel framework the spectrogram separation can outperform the time-domain solution (considering that phase component is not updated through the spectrogram separation pipeline and the phase of the mixed signal was used in the reconstruction layer).
The proposed multi-channel framework shown to be effective.
Scores were reported for multiple angle differences between the speakers; in single-channel framework results did not change significantly over the angle difference between the speakers. However, scores reported on the set of angle differences had more deviation for multi-channel setup.
The stability of proposed speech separation models on different configurations has been studied.
The scores confirm that, when enough data is available to train the models the spectrogram separation provides a more stable performance. Waveform separation requires more parameter tuning, but it is advantageous concerning the network receptive field.
Considering the fact that frequency-domain analysis can be performed within a large window such as $L=512$, it provides more efficient solution in terms of memory-usage.

\section{Conclusion}
\label{sec:con}

In this paper, we presented a unified framework for speech separation on 2-mixed speakers. 
The unified framework incorporates single-channel and multi-channel structures, and also includes the pipeline for both frequency-domain and time-domain input representations.
Through our proposed end-to-end unified framework, we improved performance of the state-of-the-art single-channel solution in frequency-domain. 
Multi-channel structure using spatial location information of the sources outperformed the single-channel framework, consistently. 
In addition, a new method proposed to control the latency with a negligible degradation in speech separation performance. 
To evaluate our proposed solutions, we simulated spatialized reverberant WSJ0-2mix and LibriSpeech-2mix.
These two corpora were simulated differently which helped to evaluate the robustness of the solutions more accurately. 
Results were reported on a set of angle differences to emphasize the strengths and weaknesses of each pipeline in the unified framework.
The experimental results confirm that overall, spectrogram-based speech separation can achieve competitive (and in some cases even better) performance for both single-channel and multi-channel inputs comparing against the waveform-based speech separation. In addition, it achieves more stable performance over different network structures, and facilitates tuning the hyper-parameters.
Spectrogram-based speech separation is also shown to be more efficient in terms of memory-usage.
The PESQ scores also were reported that confirm the overall observation from the scores achieved by SDR and Si-SNR.

In future, the spectrogram-based speech separation can be improved further with using either the complex-valued inputs to the model, or with passing a better phase component to the reconstruction layer. The performance and latency can improve further as well with incorporating a better separator neural network.

% or


%\appendices
%\section{}

\section*{Acknowledgment}
The authors would like to thank Jian Wu of the Northwestern Polytechnical University, Xi'an, China for very helpful discussions.

\ifCLASSOPTIONcaptionsoff
  \newpage
\fi



\newpage
\bibliographystyle{IEEEtran}
\bibliography{mybib}
\end{document}